\documentclass[letterpaper, 10 pt, conference]{ieeeconf}  

\IEEEoverridecommandlockouts                              

\usepackage{amssymb}  
\usepackage{graphics} 
\usepackage{graphicx}
\usepackage[tight,footnotesize]{subfigure}
\long\def\invis#1{}
\usepackage{cite}
\usepackage{url}
\usepackage{hyphenat}
\usepackage{amsmath}
\usepackage{xcolor}
\usepackage{siunitx}

\newcommand\sect[1]{Section~\ref{#1}}
\newcommand\fig[1]{Figure~\ref{#1}}
\newcommand\tab[1]{Table~\ref{#1}}

\newcommand\alg[1]{Algorithm~\ref{#1}}
\newcommand\etal{\textit{et al.\ }}

\usepackage{algpseudocode}
\usepackage{algorithm}

\newcommand\acomment[1]{\textcolor{purple}{A:#1}}

\title{\LARGE \bf
Multi-robot Dubins Coverage with Autonomous Surface Vehicles 
}

\author{Nare Karapetyan, Jason Moulton, Jeremy S. Lewis, Alberto Quattrini Li, Jason M. O'Kane, and Ioannis Rekleitis
\thanks{The  authors are with the Computer Science and Engineering Department, University of South Carolina, Columbia, SC, USA
        {\tt\small [nare,moulton]@email.sc.edu,  [lewisjs4,albertoq,jokane,yiannisr]@cse.sc.edu}}%
}
\begin{document}

\maketitle
\thispagestyle{empty}
\pagestyle{empty}

\begin{abstract}
In large scale coverage operations, such as marine exploration or aerial monitoring, single robot approaches are not ideal, as they may take too long to cover a large area. In such scenarios, multi-robot approaches are preferable. Furthermore, several real world vehicles are non\hyp holonomic, but can be modeled using Dubins vehicle kinematics. This paper focuses on environmental monitoring of aquatic environments using Autonomous Surface Vehicles (ASVs). In particular, we propose a novel approach for solving the problem of complete coverage of a known environment by a multi-robot team consisting of Dubins vehicles. It is worth noting that both multi-robot coverage and Dubins vehicle coverage are NP-complete problems. As such, we present two heuristics methods based on a variant of the traveling salesman problem---$k$-TSP---formulation and clustering algorithms that efficiently solve the problem. The proposed methods are tested both in simulations to assess their scalability and with a team of ASVs operating on a 200 $\textbf{km}^2$ lake to ensure their applicability in real world.

\end{abstract}

\section{INTRODUCTION}
This paper addresses the problem of covering a large area for environmental monitoring with multiple Dubins vehicles.
Coverage is a task common to a variety of fields. The application areas can be classified based on the scale of operations,  by the necessity to ensure the coverage of all available free space (termed \emph{complete coverage}), and by whether there is prior knowledge of the environment.  
From small scale household tasks such as vacuum cleaning and lawn mowing to large scale operations such as automation in agriculture, search and rescue, environmental monitoring, and humanitarian de\hyp mining, coverage is a key component. See \cite{Choset-2001, Galceran-2013} for in\hyp depth surveys. Finding a solution to the \emph{coverage problem} means planning a trajectory for a mobile robot in a way that an end\hyp effector, often times the body of the robot, passes over every point in the available free space. Employing multiple robots can reduce the coverage time cost, and, in hazardous conditions, such as humanitarian de\hyp mining, increase the robustness by completing the task even in the event of accidental ``robot deaths.'' The use of multiple robots  however, increases the logistical management and the algorithmic complexity. 

Covering an unknown environment, termed \emph{online coverage}~\cite{Rekleitis-2008},  focuses on ensuring that no part is left uncovered and on minimizing repeat coverage. In contrast, when covering a known environment, the focus is on performing the task as efficiently as possible~\cite{Xu_et_al.-2014}. As mentioned above, another classification is between ensuring complete coverage versus, in limited time, ensuring that the most interesting areas are covered~\cite{RekleitisCRV2017}. Furthermore, the scale of the environment in conjunction with the speed and endurance of the robot(s) classify the coverage task as small, medium, or large scale. For example, a flying vehicle with \SI{30}{\minute} battery life and  an average speed of \SI[per-mode=symbol]{40}{\km\per\hour} can cover a trajectory of \SI{20}{\km}, while an autonomous surface vehicle (ASV), moving at \SI{10}{\knot} (\SI[per-mode=symbol]{5}{\m\per\second}) for five hours will travel approximately \SI{90}{\km}. 

\begin{figure}[t]
  \centering
  \fbox{\includegraphics[width=0.43\textwidth]{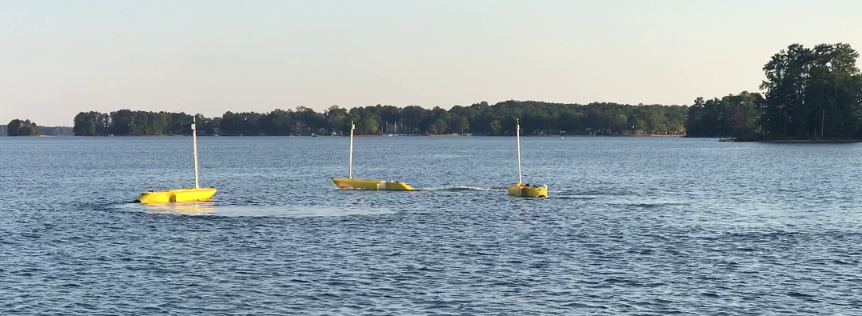}}
  \caption{Three autonomous surface vehicles during coverage experiments at Lake Murray, SC, USA.}
  \label{fig:beauty}
\end{figure}

In this paper we focus on the monitoring of aquatic environments. The vehicles
of choice are ASVs that were custom-made at the University of South Carolina (see \fig{fig:beauty}).
Aquatic environments, in general, require large scale operations. For
example, one of the testing grounds used---Lake Murray---has a surface of over
\SI{200}{\square\km}. Many ASVs, similar to fixed wing aircraft, are governed by
Dubins vehicle kinematics~\cite{Dubins}; i.e., Dubins vehicles cannot turn in
place. More formally, a Dubins vehicle is defined as a vehicle which may only
follow line segments and arcs with radius greater than some specified minimum
with non-negative velocity, i.e., they may not back up.
Recent work~\cite{7392731,RekleitisIROS2017a}
presented an efficient approach to cover an area by a single Dubins vehicle. We
extend the proposed algorithm to multiple robots based on recent work on
multi\hyp robot coverage~\cite{RekleitisIROS2017b} for holonomic robots. This
work ensures a more efficient division of labor between robots, particularly for large
scale environments. Efficiency is measured as a combination of the utilization of the robots and
the reduction of the maximum coverage cost.  The idea is that robots are limited battery life; as such, the workload should be evenly distributed.
We present two methods. In the first one, the efficient path
produced by the algorithm proposed by Lewis \etal \cite{RekleitisIROS2017a} is divided to
approximately equal parts, in terms of path length,  and each part is assigned to a
different robot. In the second method, the target area is divided into equal parts, based on
the team size, and then the single robot algorithm~\cite{RekleitisIROS2017a} is
applied to each area.

Experimental results from several simulated experiments show that indeed the utilization of the
robots is maximized and the maximum coverage cost is minimized. Moreover, the approach is 
scalable to a large number of robots.  Field trials with a single robot, 
a team of two, and a team of three ASVs, demonstrated the feasibility of the proposed
approach with real robots executing plans generated by the planner and highlighted several practical challenges.   

The next section discusses related work for the complete coverage problem using
either single or multi\hyp robot systems.  \sect{ch:probFormul}  presents the
problem statement and, in the following section, an outline of the proposed
approach is discussed.  The  experimental setup is presented in
\sect{sec:results} together with results from simulation and from the
deployment of a team of ASVs in Lake Murray, SC, USA. Finally, a discussion of
lessons learned together with future directions of this work concludes the
paper in \sect{sec:conc}.

\section{RELATED WORK}
\label{ch:litReview}

There are numerous ways to formulate coverage, including \emph{static} or
\emph{dynamic} coverage, \emph{complete} or \emph{partial}, \emph{offline}  or \emph{online} \cite{Choset-2001,Galceran-2013}. In addition, there are
many different approaches to tackle such a problem, such as defining it as
graph partitioning problem, performing region-based decomposition, or defining
it as sub-modular optimization problem \cite{7353766,Galceran-2013}.

When prior information about the environment is available as a map, the
coverage is called offline \cite{Choset-2001}.  One of the approaches widely
used in offline coverage algorithms is based on area decomposition.
Choset \cite{Choset-2000} proposed a cellular decomposition technique, called
\emph{boustrophedon decomposition} (BCD). In his work, the coverage are is decomposed into obstacle free cell. A lawnmower pattern is typically executed to cover each cell. Other approaches were also used for decomposing areas based on
Morse decomposition \cite{Acar2002J2} or grid-based decomposition \cite{Gabriely-2001}. 

 Some of the  grid-based methods to single robot coverage were adapted for multi\hyp robot systems as
 well~\cite{Hazon-MSTC,agmon2008giving, Faziletal.2010, 6608838}.  The
 robustness and efficiency of the  systems proposed by that body of work depend
 on the resolution of the input representation. Because the size of each cell is
 typically based on the size of the sensor footprint, the coverage becomes more
 challenging in environments with many obstacles, as the footprint size increases.

Polynomial time algorithms were proposed for solving
single robot coverage using a boustrophedon decomposition based
approach~\cite{Rekleitis2010a, Xu_et_al.-2014}.  In contrast to the original
algorithm, in these approaches, the problem is represented as the \emph{Chinese postman problem (CPP)}.
The latter is a graph routing problem, of finding a minimum\hyp cost closed tour that
visits each edge at least once. Edmonds and Johnson~\cite{cpp} 
found a polynomial\hyp time solution for CPP.

When considering the coverage problem for robots with turning constraints, a
simple boustrophedon coverage plan may introduce wasted time---that is, time
spent out of the region of interest because of the constraints and thus not actually covering. The Dubins vehicle is a common robot model
in coverage problems, and Savla, Bullo, and Frazzoli~\cite{SavBulFra07} consider a
control-theoretic solution. In our work, however, we provide an algorithmic
approach to minimizing the path length by minimizing the time spent not actively
covering because of the motion constraints.

Reducing traversal time by considering motion constraints is not a new idea
in coverage. Both Huang~\cite{Huang2001} and Yao~\cite{Yao2006} minimize the path
length by using motion constraints in their environmental decompositions. Both of them, however, seek to reduce the amount of rotation required by the robot,
while we optimize the solution by carefully selecting how the robot transitions from covering to not-covering. 

This idea is related to the traveling salesman problem (TSP) with Dubins curve
constraints, called Dubins traveling salesman problem. 
In \cite{NyFerFra12,SavFraBul05}, 
the Dubins TSP is defined as metric TSP with the additional constraint that paths
between nodes must adhere to a minimum turn radius necessary for the covering
vehicle's transition between nodes.

Some of the presented methods based on cellular decomposition  were also designed for multi-robot systems~\cite{Rekleitis-2008}, assuming restricted communication. Avellar \etal\cite{s151127783} present a
multi\hyp robot coverage approach that operates in two phases: 
decomposing the area into line\hyp sweeping rows, based on which a complete graph is constructed to be used in the second phase, where the vehicle routing problem~\cite{Toth:2001:VRP} is solved. Field trials with  Unmanned
Aerial Vehicles (UAV) showed that their proposed approach provides minimum-time coverage.
However, that algorithm is only applicable for obstacle free environments.

In our previous work~\cite{RekleitisIROS2017b}, we presented a
communication-less multi-robot coverage algorithm based on efficient single robot coverage. Even if the proposed methods demonstrate better performance on
robots utilization and almost optimal work division, the robots utilization is
dependent on the number of obstacles in the area. As in that work clustering is
based on boustrophedon cells, a small number of obstacles will result in small number of cells, and consequently less clusters per regions. Note that, however, the solution generated did not take into account any kinematic constraints of the robots.

A large body of work in multi-robot systems assumes that there 
is some form of communication between the robots 
\cite{xin2016coordinated}. Some of them came up with 
alternative implicit communication means, such as trail of 
other robots \cite{wagner1999distributed, Janani:2016:Abst, Janani2016}. 
Nevertheless, this type of communication is impractical in 
aquatic or aerial environments.

The graph routing problems such as TSP and CPP have also their definition for
multiple routes: finding $k$ routes that visit non overlapping vertices of
the graph, such that the union of those clusters are the exact set of 
vertices in the TSP case. This problem is called $k$\hyp TSP problem. When
edges are considered instead of vertices, the problem is called $k$\hyp CPP. Both these
problems and their variations were shown to be NP-complete
\cite{Frederickson:1976}.  

In this paper, differently from the current state of the art, we address multi-robot coverage for Dubins vehicles, for which no solution is readily available. In the following section, the problem is formally defined.

\section{PROBLEM STATEMENT}
\label{ch:probFormul}

The \emph{Dubins multi-robot coverage problem} can be formulated as follows. We assume to have $k$ homogeneous robots, with no communication capabilities, equipped with a sensor with fixed-size footprint $s$, and with Dubins constraints---namely, the robots have a minimum turning radius $r$, that constrains the robots to follow line segments and arcs with radius greater than $r$, and they cannot drive in reverse.
Such robots are deployed in a $2$D-bounded area of interest region $\mathcal{E} \subset \mathbb{R}^2$. The objective is to find a path $\pi_i$ for each robot $i$, with $1\leq i \leq k$, so that every point in the region of interest $\mathcal{E}$ is covered by at least a robot's sensor.

An efficient solution is one that minimizes the length of the trajectories for the robots, while at the same time ensuring that the workload on the robots are evenly distributed. This is motivated by the fact that homogeneous robots have the same limited battery life, and thus, to cover a big region, it is better to utilize all of them for the coverage task.

In practice, this means that an efficient algorithm finds $k$ non-overlapping
regions $\mathcal{E}_i \subset \mathcal{E}$ such that $\mathcal{E} =
\bigcup_{1}^{k}(\mathcal{E}_i)$, where each robot $i$ can perform a calculated covering trajectories $\pi_i$.
Note that $\pi_i$ includes the whole path robot has to follow: a robot starts from an initial starting point $v_s$, goes to a point of entry to a partition of the interest region $\mathcal{E}_i$, covers fully $\mathcal{E}_i$, and goes back to $v_s$. We call the coverage cost---i.e., the traveled distance---of a single robot covering
$\mathcal{E}_i$ as $c(\mathcal{E}_i)$. As such, we can define the optimization problem of Dubins multi-robot coverage as a \textit{MinMax} problem: minimizing the maximum cost $\max_{1}^{k}(c(\mathcal{E}_i))$ over all robots.

\section{PROPOSED METHODS}
\label{sec:methods}

In this section we introduce the terminology used in the subsequent sections.
Next, we present the two $k$\hyp coverage algorithms. The first algorithm
builds an optimal tour and splits it between multiple robots and is called
\textit{Dubins Coverage with Route Clustering (DCRC)}. The second algorithm
first divides the area between robots and then starts route planning, which we
refer to as \textit{Dubins Coverage with Area Clustering (DCAC)}.

\subsection{Terminology}
\label{sec:term}


A \emph{cell} is defined as a continuous region containing only the area of interest that one of the robots must
cover entirely. The cells are the result of a BCD decomposition
\cite{Choset-2000}. The Dubins coverage algorithm by Lewis \etal \cite{RekleitisIROS2017a}---referred to as Dubins coverage solver (DCS)---is
	a process by which a coverage problem is mapped to a graph for which a solution
	to the TSP results in a single coverage path.

	The DCS algorithm divides cells into a collection of \emph{passes}, defined
	as the smallest unit of coverage; each of which is axis-aligned and has a width
	equal to the robot's sensor footprint. Each pass becomes the node of a directed, weighted Dubins graph
	$G_d = (E_d, V_d)$. The edges of $G_d$  are defined as the Dubins path from a
	source node to a target node. The weight of an edge $w(u,v)$ is then the length of the
	segments and arcs of the Dubins path between two passes $u$ and $v$.  The output of the DCS
	algorithm is an optimal \emph{Hamiltonian path} $R = \{v_1, v_2,..., v_n\}$, where $v_i
	\in V_d$ and $n$ is the number of passes, that is $|V_d|$.

		\subsection{Dubins Coverage with Route Clustering (DCRC)}
		\label{sec:r1c2}

		Our first approach for multi-robot Dubins coverage is based on DCS  
		and Coverage with Route Clustering (CRC) method \cite{RekleitisIROS2017a}.

		The CRC algorithm creates cells applying the BCD algorithm on a binary image 
		of the area with obstacles\cite{RekleitisIROS2017b}. 
		Then, boustrophedon cells are turned into edges of a weighted graph---called Reeb Graph---on which $k$-Chinese Postmen Problem ($k$-CPP) is solved.
		The result is a $k$-partitions of an optimal route. 

		To address Dubins constraint in this paper we are interested in solving the $k$-TSP problem instead of $k$-CPP. The pseudocode for DCRC is presented in \alg{alg:dcrc}.
		Line 1 gets an optimal Hamiltonian path $R=\{v_1, v_2, ..., v_n\}$, where the vertices are passes, by using the DCS algorithm to solve the single-robot Dubins Coverage problem with the DCS algorithm.
		Its cost $c(R)$ is given by the initial traveled distance to get to the region of interest $c(v_s,v_1)$, the sum of the costs $w(v_{j}, v_{j+1})$ to cover passes $v_{j}$,$v_{j+1}$, and the cost  $c(v_{n}, v_{s})$to go back  to the starting point $v_{s}$ (Line 3).
		Note that the travel cost  $c(v, u)$ is defined as  Euclidean distance between midpoint coordinates of corresponding $u$ and $v$ passes.
		The resulting optimal path $R$ is split into $k$ subtours $\{R_1, R_2, ..., R_k\}$ (Lines 4\hyp7). 
		For a given starting point $v_s$, the cost of any tour $R_i=\{v_{i_1}, v_{i_2}, ..., v_{i_m}\}$ is defined as the cost of traveling from the starting point to reach a designated coverage cell, the actual cost of covering that cell and the cost of traveling back to the starting point (Line 8, where $m$ is the index of the last pass/vertex in the path).
		 Cost ${c_{\max}}$ is calculated to balance travel and coverage costs between robots (Line 3).	Such a clustering procedure was proposed in the $k$-TSP solver by Frederickson \etal \cite{Frederickson:1976}.

		\begin{algorithm}
		\caption{DCRC}
		\label{alg:dcrc}
		\textbf{Input:} number of robots $k$, binary map of area $M$,
		\hspace*{1.20cm}turning radius $r$, sensor footprint $s$\\
			\textbf{Output:} $k$ tours, 1 for each robot
			\begin{algorithmic}[1]
			\State $\textit{R} \gets \textrm{DCS}(M, s, r)$
			\State \textit{initialize for each $i$ in $k$ empty tours $R_i$}
			\State $c(R)=c(v_s,v_{j}) + \sum_{j=1}^{n-1}{w(v_{j}, v_{j+1})} + c(v_{n}, v_s)$ 
			
			\State $c_{\max}=\max\limits_{1 \leq i \leq n}\{c(v_{1}, v_i) + w(v_{i}, v_{i+1}) + c(v_{i+1}, v_1)\}$ 
			\For{\textbf{each} $\textit{i} \in \textit{1, ..., k}$}
			\While{$c(R_i) <= (c(R)-2c_{\max})*i/k + c_{\max}$}
			\State \textit{$include~next~vertex~ v ~ along~ R~ into~ R_i$}
			\State $c(R_i)=c(v_s,v_{i_1}) + \sum_{j=1}^{m-1}{w(v_{i_j}, v_{i_{j+1}})} + c(v_{i_m}, v_s)$
			\EndWhile
			\EndFor
			\end{algorithmic}

			\end{algorithm}

			The complexity of this algorithm is exponential as DCS uses an exact TSP solver.
			\invis{\acomment{We do not. We use an optimizer, though its arguable that we
				would use an exact solver if one was available and bug-free.} } The FHK
				algorithm is proved to have an approximation factor of $\frac{5}{2} -
				\frac{1}{k}$ \cite{Frederickson:1976}.

\subsection{Dubins Coverage with Area Clustering (DCAC)}
\label{sec:c1r2}

The DCAC algorithm, similar to the CAC algorithm \cite{RekleitisIROS2017b}, performs clustering of the
region of interest $\mathcal{E}$ and then finds the optimal route for each robot.
An overview of the DCAC algorithm is presented in \alg{alg:dcac}. 

In particular, the BCD algorithm is applied to decompose the environment into cells, consisting entirely of areas which should be covered (Line 1). 
Then, each cell is divided into passes (Line 2). A
corresponding graph is created from these passes (Line 3). The graph is an undirected weighted graph $G = (V, E)$,
where each vertex is located at the center of a pass;   vertices $(v_i, v_j)$ in this
graph are connected with an edge $e$ if and only if their corresponding passes
share a common edge. The cost $c(e)$ of each edge $e=(v_i,v_j)$ is defined as the Euclidean
distance between midpoints of passes. The vertices of graph $G$ are clustered performing a breadth-first search (BFS) clustering  (Line 4). 
The size of a cluster $C=\{v_1, v_2, ..., v_m\}$ is defined as $c(C)=\sum_{\{e \mid e=(v_i, v_{i+1}), 1 <  i\le m\} }c(e) $.
DCS is then applied on each resulting cluster of passes (Lines 5-7).

The clustering step in the CAC algorithm \cite{RekleitisIROS2017b} ensures that the cost of reaching the region of interest and the 
actual coverage cost per region are balanced, by assigning more passes to cover to robots that are closer to the region of interest; while the robots that have to travel more to reach the coverage area will have less passes to cover.

\begin{algorithm}
\caption{DCAC}
\label{alg:dcac}
\textbf{Input:} number of robots $k$, binary map of area $M$,
	\hspace*{1.20cm}turning radius $r$, sensor footprint $s$\\
	\textbf{Output:} $k$ tours for each robot
	\begin{algorithmic}[1]
\State $\textit{cells} \gets \textrm{BCD}(M)$
\State $\textit{passes} \gets \textrm{GenPasses}(\textit{cell}, s)$
\State $\textit{G} \gets \textrm{buildGraph}(\textit{passes},r)$
\State $\textit{C\_set} \gets \textrm{BFSClustering}(\textit{G}, k)$ \Comment{clusters of passes}
\For{$\textbf{each}~\textit{$C_i$} \in \textit{C\_set}$}
\State $\textit{tour} \gets \textrm{DCS}(\textit{$C_i$}, r, s)$
\EndFor
\end{algorithmic}
\end{algorithm}

As the complexity of TSP is exponential, by  partitioning problem into $k$ small TSP subproblems, the overall TSP performance is improved.
Nevertheless, the complexity will still remain exponential.


\section{EXPERIMENTS}
\label{sec:results}
The proposed method has been first evaluated with simulation tests for large environments
within a custom simulator that accounts for Dubins constraints, to test the optimality of such an approach and its scalability.

\begin{figure}[h]
	\centering      
	\includegraphics[width=0.6\columnwidth]{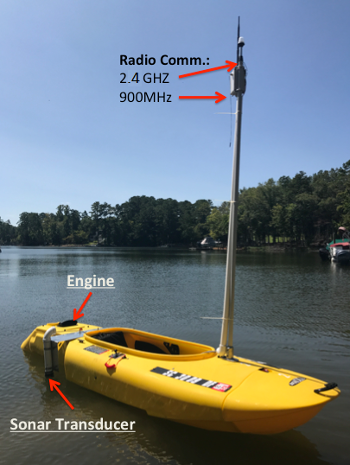}
	\caption{Experimental ASV Setup.}
	\label{fig:MokaiAn}
	\end{figure}

Second, we modified a fleet of jet-drive Mokai ES-Kape sport kayaks, shown in \fig{fig:MokaiAn}, to be autonomous surface vehicles, and used them for validating the proposed approach with real robots. The goal includes checking if the assumptions made hold in the real world. The ASVs are equipped with a SONAR transducer collecting depth measurements with a frequency of \SI{1}{\Hz}, a PixHawk controller for waypoint navigation and safety behaviors, and a Raspberry Pi with the Robot Operating System (ROS) framework~\cite{ros} to record GPS and depth data.

\invis{Adding two more ASVs and loading their mission's according to the proposed algorithms, we conducted the test on the same area in order to minimize environmental influences on their trajectories.  Two ASVs were accurately tuned and repeatedly tested prior to this experiment.  These two ASVs were also equipped with sonar depth finders, conforming to the NMEA 0183 standard, to illustrate the capability of this platform to host several different sensors. }

\invis{The goal of USC's Autonomous Field Robotics Lab was to create a low-cost platform based based on commercially available components, that can be built quickly and reliably to enable implementation of these coverage algorithms.  Additionally, the platform should be scalable to accept emerging sensor and communication capabilities, allowing for robust state estimation and multi-robot decision making.   

	The platform selected is a Mokai ES-Kape sport kayak with a jet drive, shown in \fig{fig:MokaiAn}, configured as an autonomous surface vehicle.}

	\subsection{Simulated Results}

	The simulation was performed for three large input environments maps taken from Lake
	Murray and rural Quebec area. The maps differ in terms of size and shape complexity.  We have evaluated both DCRC and DCAC algorithms with a different number of robots, that is $k \in \{1, 2, 5, 10\}$ robots. \invis{The cost of the simulated path are scaled pixel values.}
	 The
		baseline for comparing the costs of each tour is the cost of optimal route
		produced by TSP algorithm. 
		As the problem is defined as MinMax problem, we
		consider the value of the maximum cost per robot along with the ideal cost as metric. The ideal cost is defined by dividing the single optimal route cost to the number of robots. 
		Another metric considered in this paper is the \emph{robots' utilization}, that is
		the ratio between the number of robots used and the total number of robots available.
		However, in the following, results with the robots' utilization are not reported: in all the experiments, the robots' utilization is $100\%$, differently from the results obtained in~\cite{RekleitisIROS2017b}. This can be explained by the additional decomposition of the boustrophedon cells in passes, which allows the algorithms to distribute cells more evenly to robots.

\begin{figure*}[ht]
		\begin{center}
		\leavevmode
		\begin{tabular}{ccc}
		\subfigure[]{\includegraphics[width=0.2\textwidth]{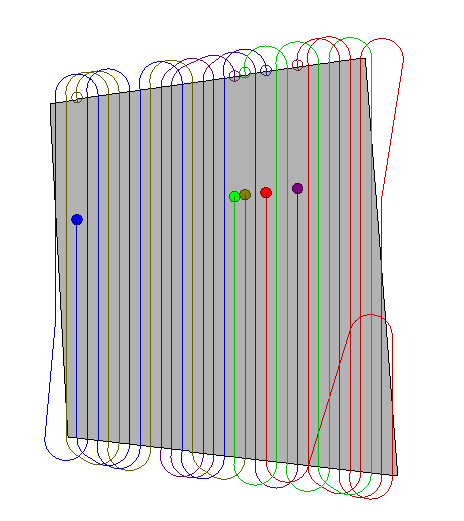}\label{fig:sima}}&
		\subfigure[]{\includegraphics[width=0.27\textwidth]{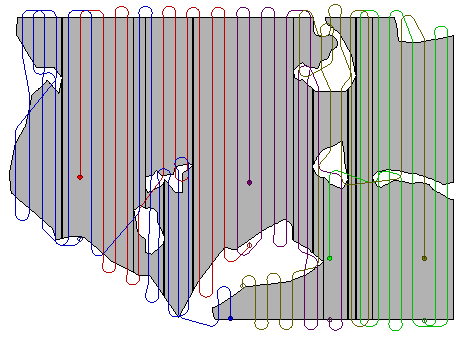}\label{fig:simb}}&
		\subfigure[]{\includegraphics[width=0.36\textwidth]{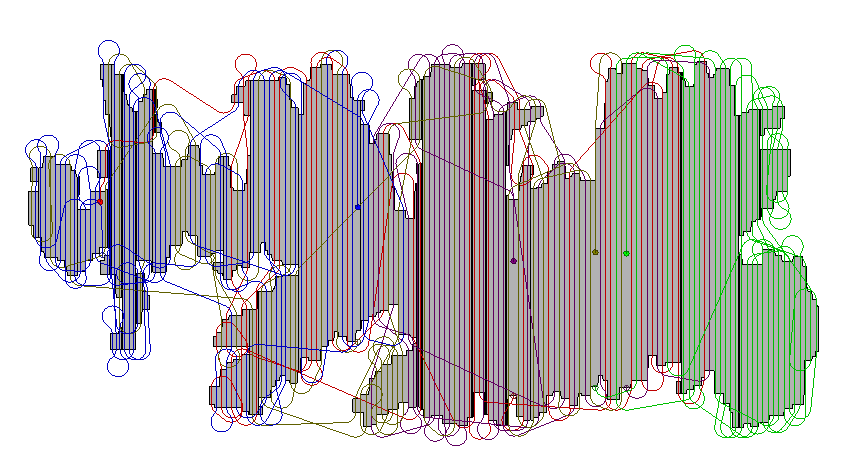}\label{fig:simc}}\\
			\subfigure[]{\includegraphics[width=0.2\textwidth]{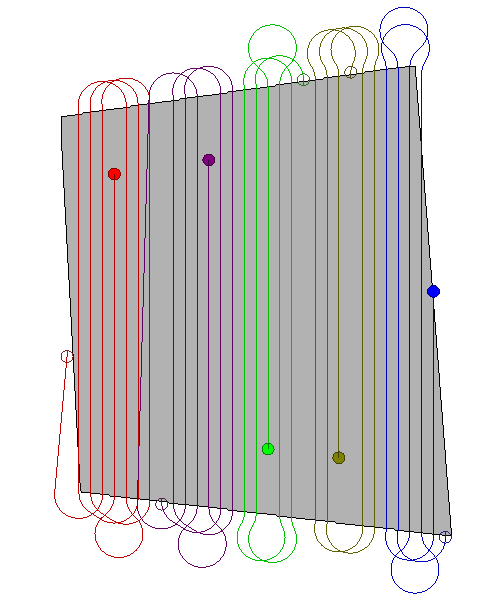}\label{fig:simd}}&
			\subfigure[]{\includegraphics[width=0.27\textwidth]{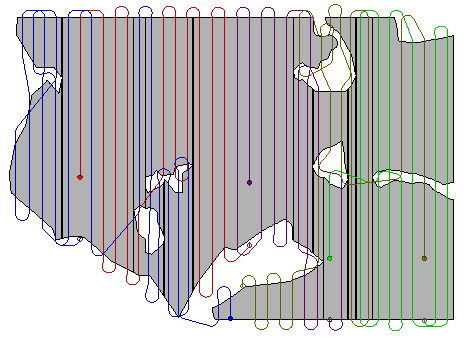}\label{fig:sime}}&
			\subfigure[]{\includegraphics[width=0.36\textwidth]{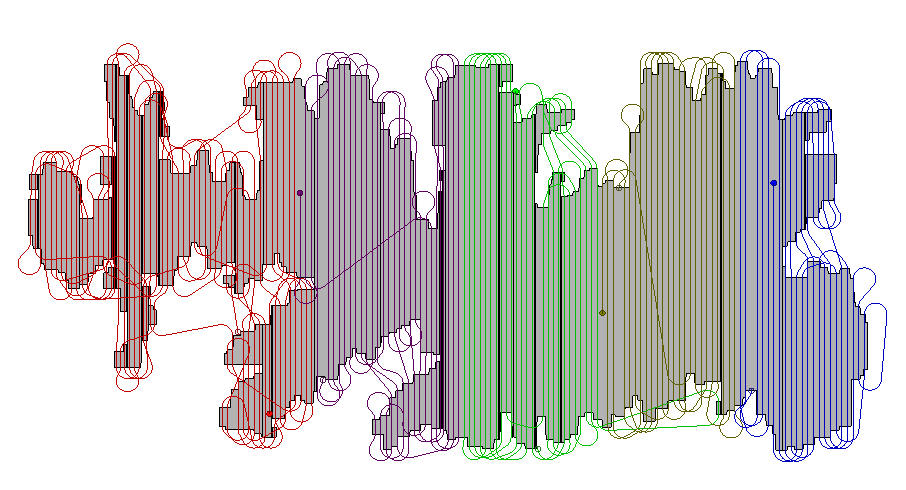}\label{fig:simf}}\\
				\end{tabular}
				\end{center}
				\caption{A simulation instance of DCRC (first row) and DCAC (second row) algorithms with 5 robots performing coverage over the area of interest indicated in gray, where the first column shows a small segment in the Lake Murray (\SI{200x200}{\metre}); second column, Rural Quebec (\SI{13 x 10}{\km}); third column, the complete Lake Murray (\SI{25x25}{\km}).} 
				\label{fig:sim}
				\end{figure*}

		\fig{fig:sim} shows the paths followed by $5$ robots on the three environments considered, using both algorithms. Qualitatively, it can be observed that DCAC produces paths where robots mostly transition to adjacent passes, while with DCRC, robots go to one pass to another that are typically not adjacent. This fact makes the robots following the paths generated by DCAC going out from the region of interest because of the minimum turning radius---compare for example \fig{fig:sim} (a) and (b). Those tighter turns contribute to an increase in the overall cost. 

\begin{figure*}[h]
				\begin{center}
				\leavevmode
				\begin{tabular}{ccc}
				\subfigure[]{\includegraphics[width=0.25\textwidth]{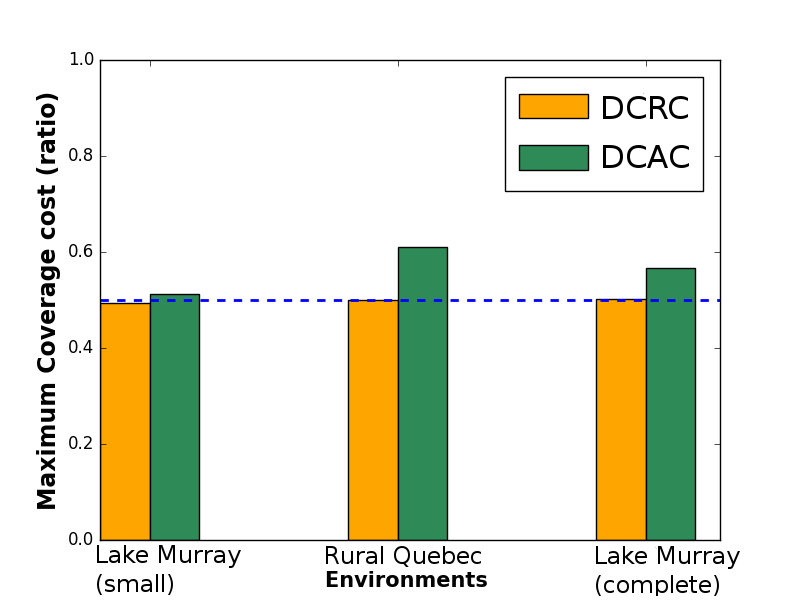}\label{fig:s1}}&
				\subfigure[]{\includegraphics[width=0.25\textwidth]{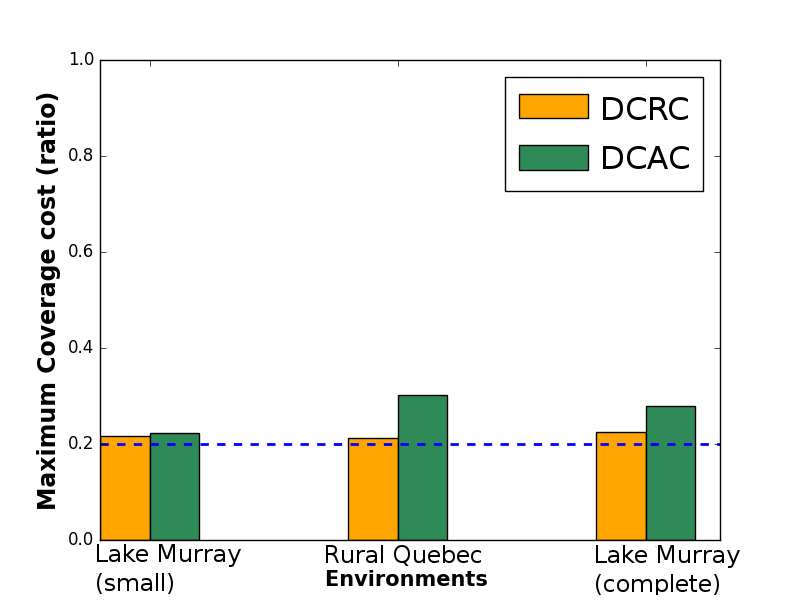}\label{fig:s2}}&
				\subfigure[]{\includegraphics[width=0.25\textwidth]{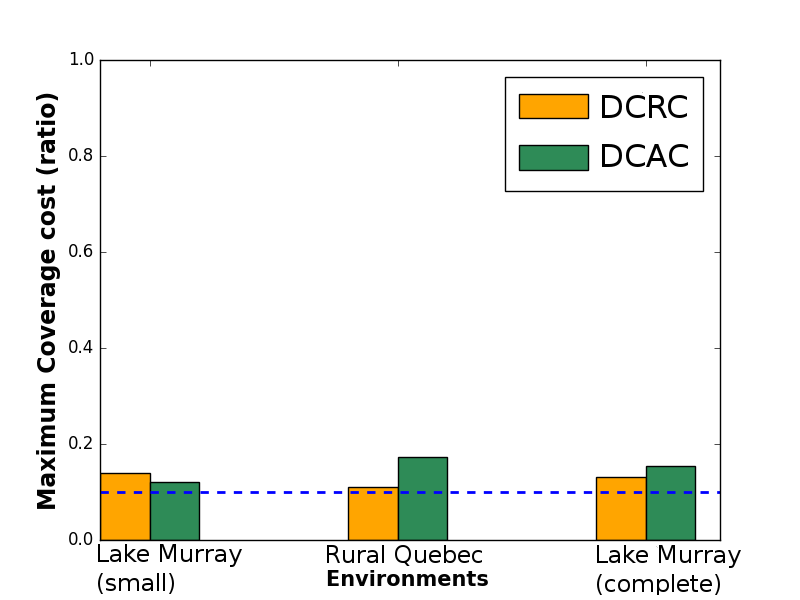}\label{fig:s3}}\\
					\end{tabular}
					\end{center}
					\caption{The comparison of actual maximum coverage cost and the ideal cost for three different environments for (a) $k=2$ (b) $k=5$ and (c) $k=10$ robots.\label{fig:simres}}
					\end{figure*}

		Indeed, as illustrated in \fig{fig:simres}---which shows the ratio between maximum coverage cost and ideal cost---DCRC has better performance. For example, in the Rural Quebec environment with $5$ robots, DCRC has maximum coverage cost ratio of $0.2$, while for DCAC is $0.3$.

					\subsection{Field trials}
					Given the better performance of DCRC, we validated the approach using DCRC with the ASVs, in a \SI{200 x 200}{\m} area in Lake Murray, SC. The sensor footprint used had \SI{4.5}{\m} and the turning radius of the ASV is \SI{5}{m}. A path, in the form of a waypoint sequence, was generated with the ASVs starting just outside the area of interest. In the following, a description of the experiments performed and the results obtained.

The main objective of the field trials was to ensure that the assumptions hold also with real robots, so that the ASVs are able to follow the trajectories generated by the proposed algorithms.

					\invis{In this experiment, we established a single robot coverage baseline using one ASV covering a \SI{200 x 200}{\m} area.  Each of the first two boats were also equipped with redundant fail-safe controls and recording accomplished by remote control override, waypoint navigation through PixHawk mission planning, and on-board Raspberry Pi running ROS and MAVROS.  The third ASV was tuned in the lab, but did not undergo the same regiment of field trial tuning as the others, nor was it equipped with sonar.}

					\subsubsection{Single Robot Coverage baseline}
					Similar to the simulation experiments, the single robot coverage for Dubins Vehicles algorithm~\cite{RekleitisIROS2017a} is used here as a baseline for comparison with the multi\hyp robot approach.\invis{; see \fig{fig:SingleASV} for the ASV in action. }

					\invis{A single ASV was utilized to perform coverage over a \SI{200 x 200}{\m} area; see \fig{fig:SingleASV} for the ASV in action.} 

\fig{fig:singleIdealPath} and \fig{fig:SingleRealCoverPath} present the ideal path and the path followed by the ASV, respectively,  as recorded GPS points overlaid on Google Maps. The depth measurements were combined using a Gaussian Process (GP) mapping technique~\cite{rasmussenGP} to reconstruct the floor map of that part of the lake---see \fig{fig:SingleRealDepth}.

\subsubsection{Multi-Robot Coverage Experiments}
A variety of experiments were performed using teams of two or three robots in
different areas of Lake Murray. 

\begin{figure*}[ht]
\begin{center}
\leavevmode
\begin{tabular}{ccc}
		\subfigure[]{\includegraphics[width=0.31\textwidth]{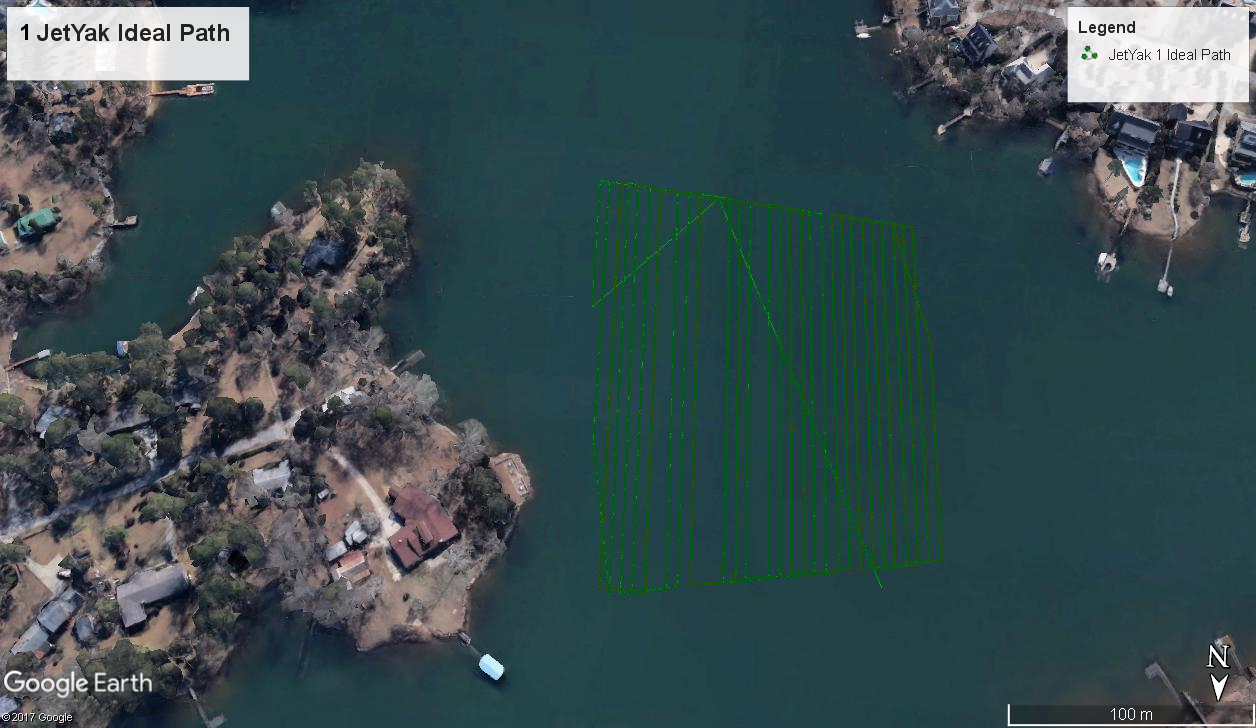}\label{fig:singleIdealPath}}&
\subfigure[]{\includegraphics[width=0.31\textwidth]{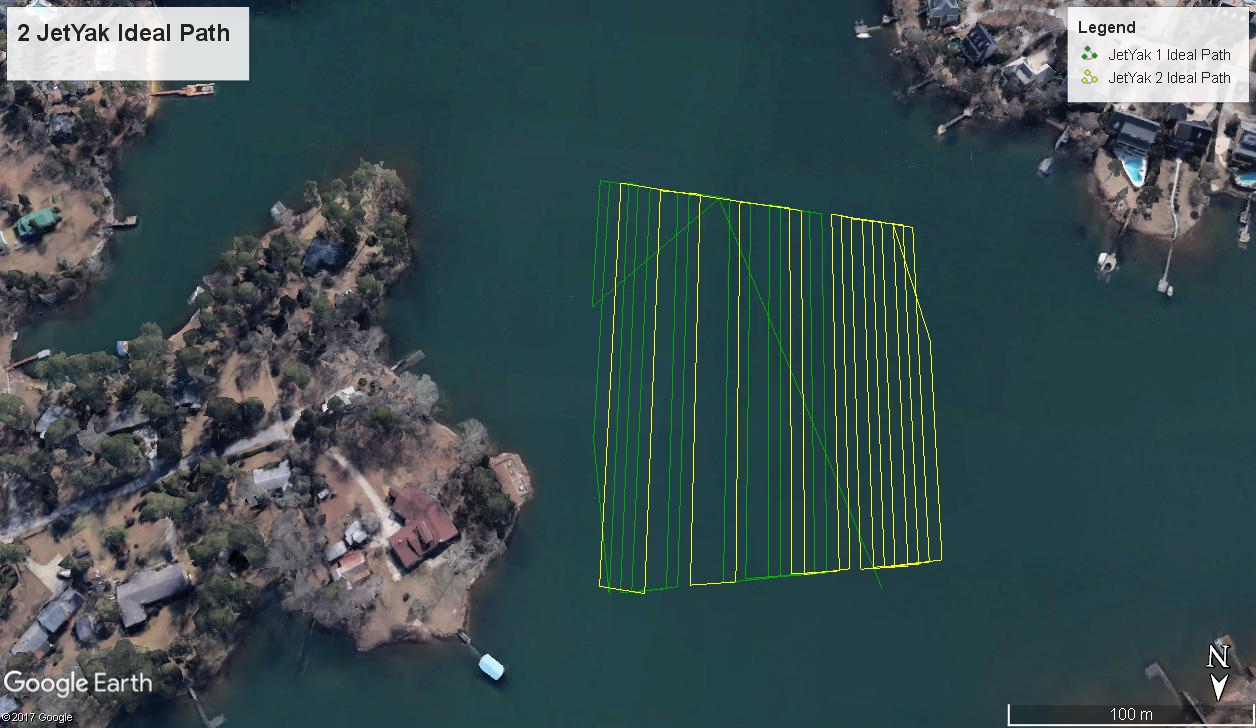}\label{fig:multi23a}}&
\subfigure[]{\includegraphics[width=0.31\textwidth]{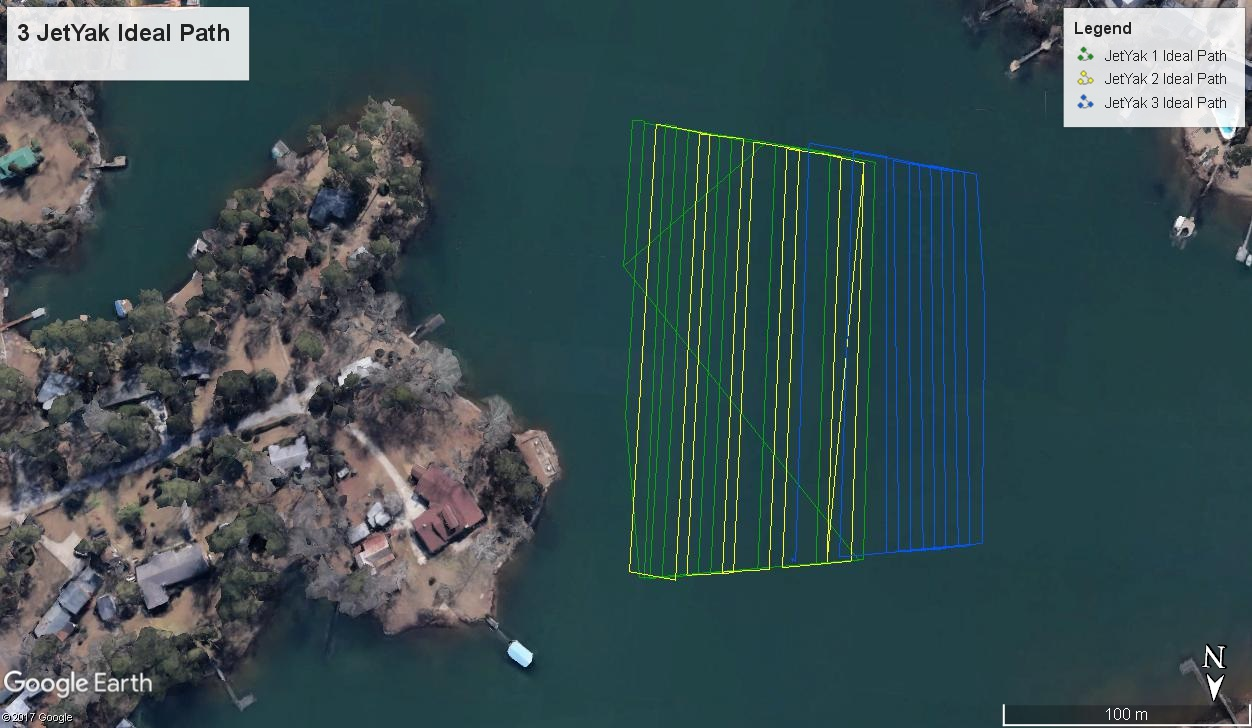}\label{fig:multi23b}}\\
\subfigure[]{\includegraphics[width=0.31\textwidth]{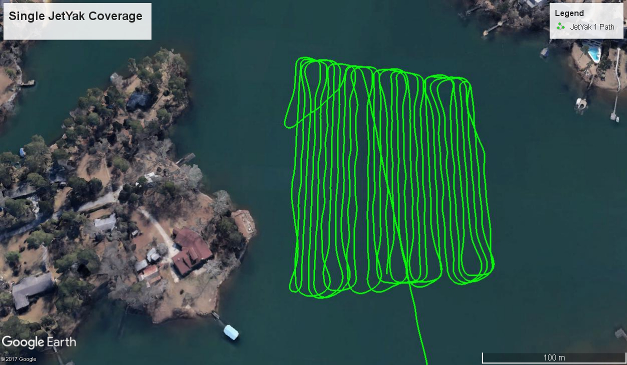}\label{fig:SingleRealCoverPath}}&
		\subfigure[]{\includegraphics[width=0.31\textwidth]{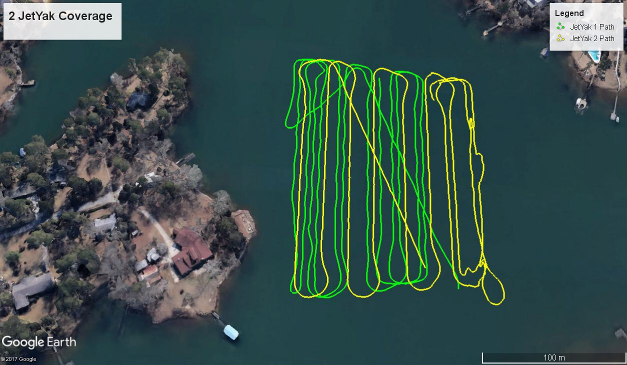}\label{fig:multi23c}}&
		\subfigure[]{\includegraphics[width=0.31\textwidth]{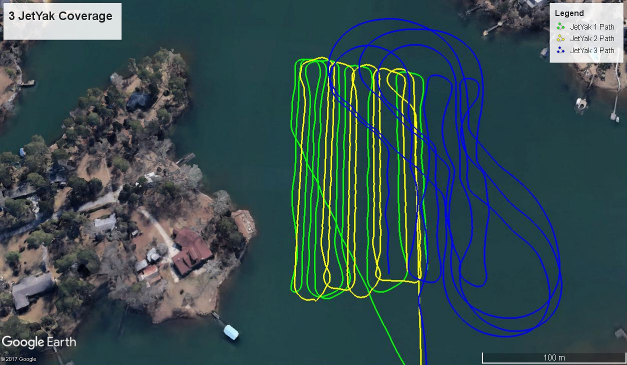}\label{fig:multi23d}}\\
		\subfigure[]{\includegraphics[width=0.31\textwidth]{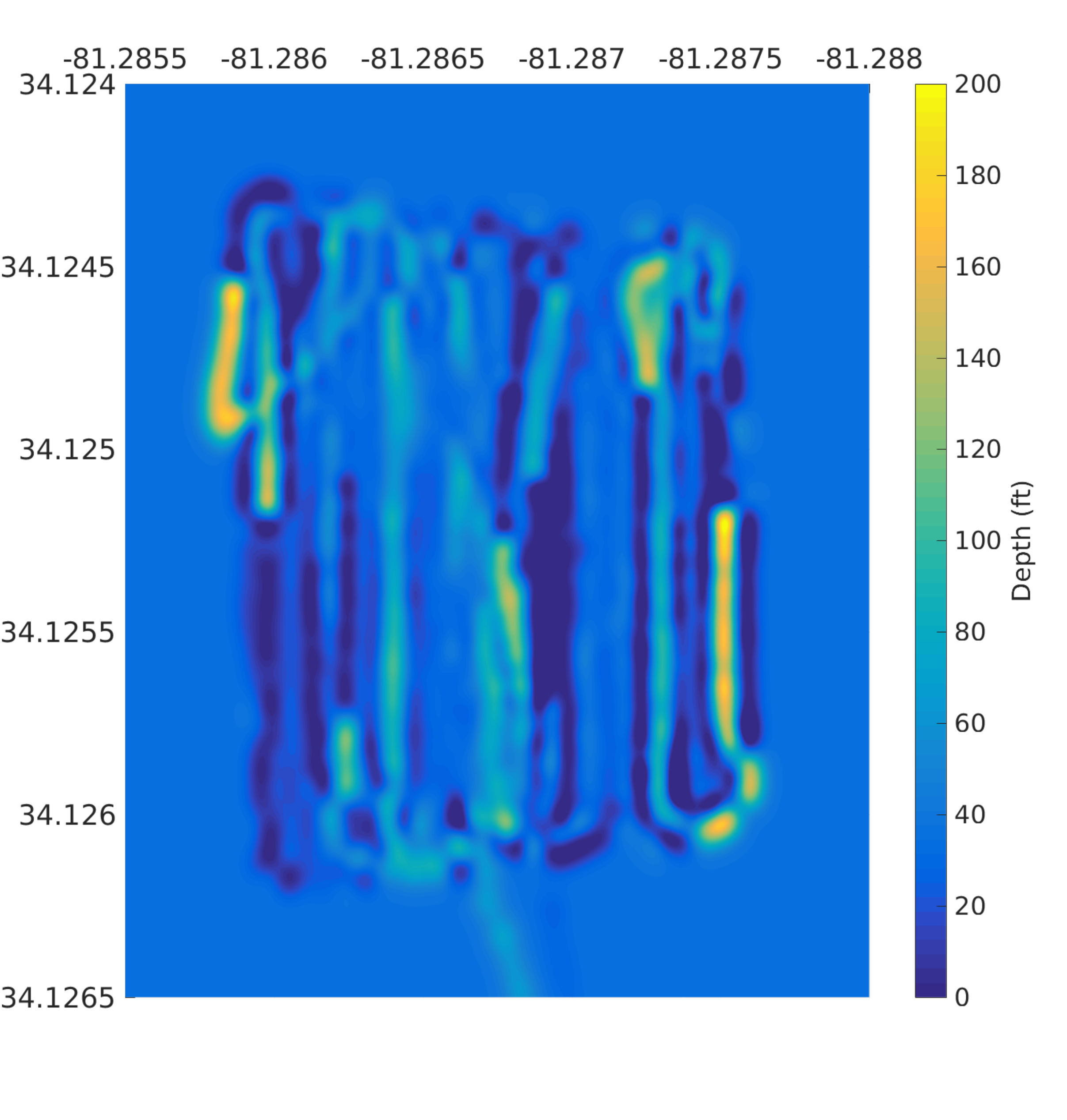}\label{fig:SingleRealDepth}}&
			\subfigure[]{\includegraphics[width=0.31\textwidth]{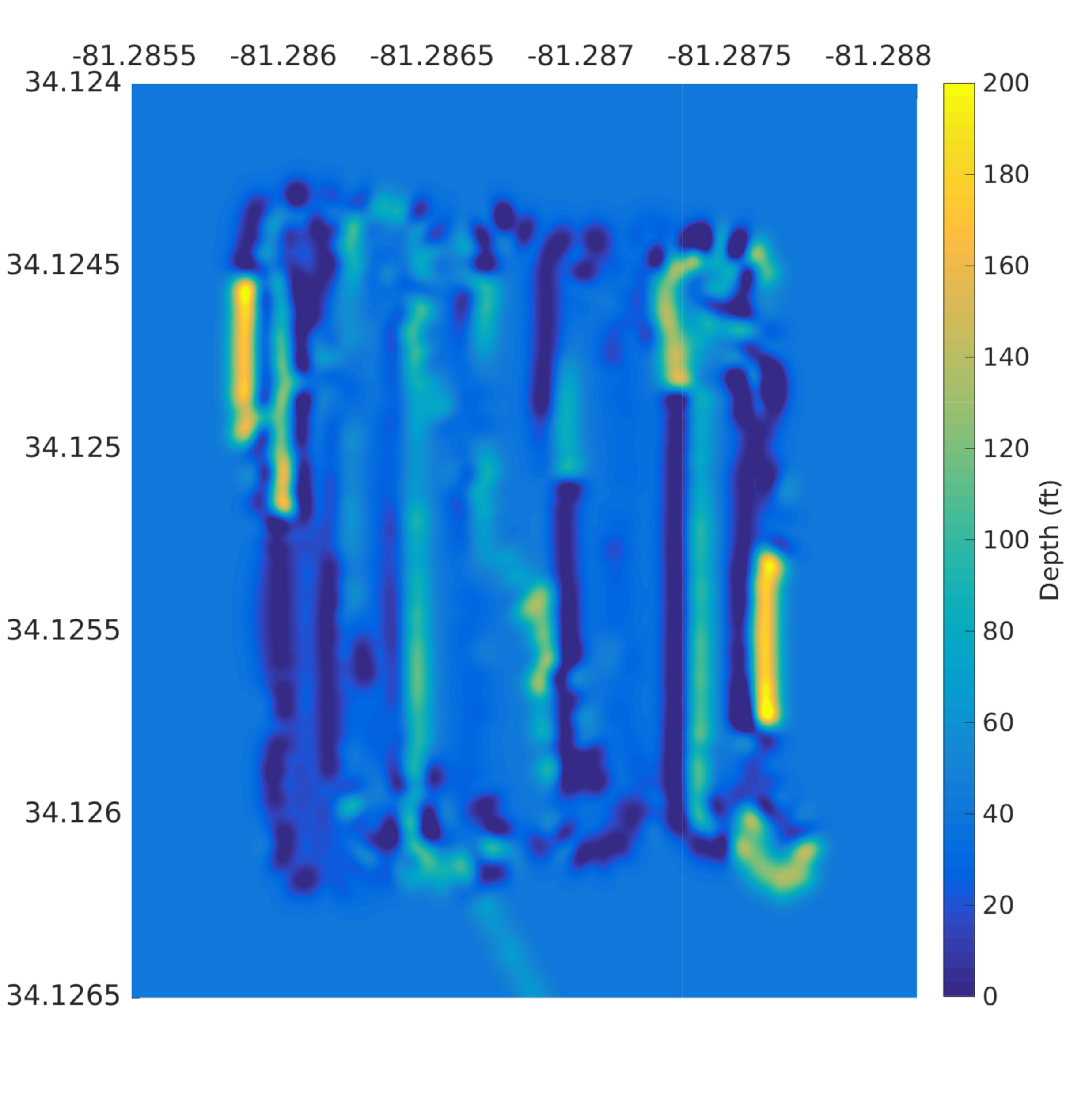}\label{fig:multi23e}}&
			\subfigure[]{\includegraphics[width=0.31\textwidth]{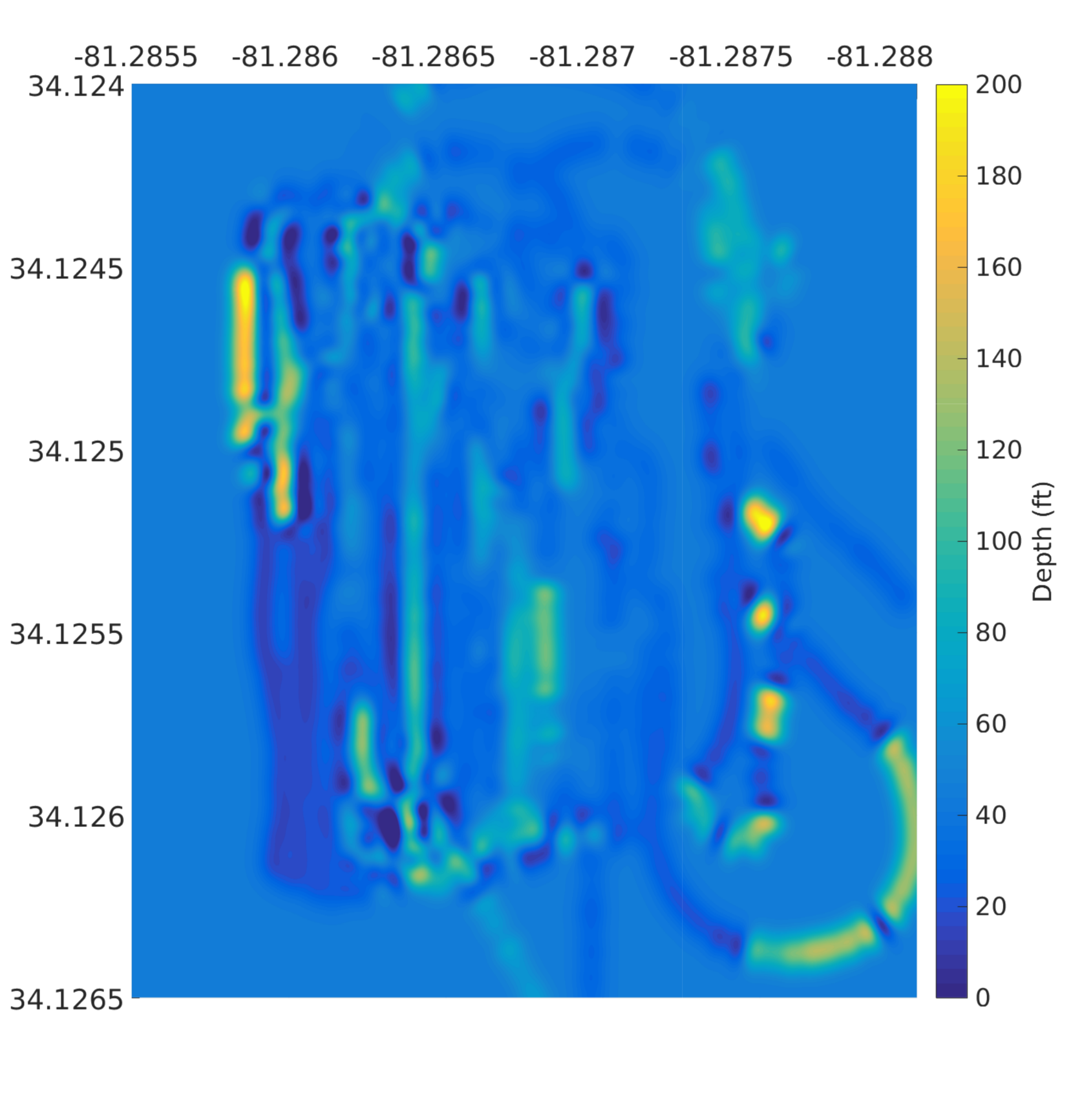}\label{fig:multi23f}}\\
				\end{tabular}
				\end{center}
				\caption{Multi\hyp robot experiments at Lake Murray, SC, USA.  \subref{fig:multi23a} Ideal path produced for two robots. \subref{fig:multi23b}  Ideal path produced for three robots. \subref{fig:multi23c} GPS track of the actual coverage path for two robots. \subref{fig:multi23d} GPS track of the actual coverage path for three robots. \subref{fig:multi23e} Depth map produced using a GP-based mapping using data from two robots.  \subref{fig:multi23f} Depth map produced using a GP-based mapping using data from three robots. }
				\label{fig:multi23}
				\end{figure*}

Figures \ref{fig:multi23a} and \ref{fig:multi23b} shows the ideal path for two and three ASVs as generated by DCRC; while Figures \ref{fig:multi23c} and \ref{fig:multi23d} shows the actual path followed by two and three robots, respectively.
As can be seen qualitatively in Figures \ref{fig:multi23a} and \ref{fig:multi23b}, the path followed by the ASVs are pretty much in line with the ideal path. The small deviations \invis{(\acomment{quantify})} are due to GPS error, current, wind, and waves from other vessels. 
As such, the proposed methods can be applied for coverage with ASVs with Dubins constraints. 

Note the ill-structured path of one of the robots (robot following the blue trajectory), result of a hardware failure and a hysteresis of its on-board PID controller. This illustrates the real world challenges with field trials: even if the boats are supposed to be identical, they are not, and they should undergo each of them an initial tuning phase of the different parameters of the boats. Such an issue opens interesting research directions on robust multi-robot coverage, including recovery mechanisms to adapt the algorithms to the new minimum turn radius and accounting for heterogeneity. \invis{However, looking at the worst case, this scenario can be addressed by our proposed approach, considering the biggest turning radius of the ASVs.}

				\invis{Each of the first two boats were also equipped with redundant fail-safe controls and recording accomplished by remote control override, waypoint navigation through PixHawk mission planning, and on-board Raspberry Pi running ROS and MAVROS.  The third ASV was tuned in the lab, but did not undergo the same regiment of field trial tuning as the others, nor was it equipped with sonar.}

The resulting multi-robot coverage is also comparable to the single-robot coverage trajectory, where only small areas were left uncovered. Indeed, the bathymetric maps resulting from the single and multi-robot coverage are similar.

The maximum traveled distances per experiment with  
				different number of robots are presented in \tab{tab2} along with the ideal traveled distance. As in case of the simulation, the ideal path length is the size of the sub area if the tasks were exactly divided to equal parts.

				\begin{table}[htbp]
				\caption{The maximum distance traveled per robot and the  cost of perfect division for multi\hyp robot coverage experiments with the real ASVs.}
				\begin{center}
				\begin{tabular}{|l|c|c|c|}
				\hline
				Number of Robots& 1  & 2 & 3   \\
					\hline
					Max Distance& \SI{6863}{\metre} & \SI{2905}{\metre} & \SI{3356}{\metre} \\
						\hline
						Ideal Distance& \SI{6863}{\metre} & \SI{3431.5}{\metre} & \SI{2287.7}{\metre}  \\
							\hline
							\end{tabular}
							\end{center}
							\label{tab2}
							\end{table}%

\section{CONCLUSIONS}
\label{sec:conc}
This paper presented a novel approach for multi\hyp robot coverage utilizing multiple ASVs governed by Dubins vehicle kinematics. 
Both presented algorithms are extending our previous work on efficient multi-robot coverage with Dubins coverage algorithm. 

The further clustering of the area ensures 100\% utilization of robots. 
We show the validity and scalability of both approaches in simulation. 
The experiments show that both algorithms result in almost optimal solutions. 
Nevertheless, DCRC algorithm demonstrated slight advantage over DCAC algorithm in terms
of coverage cost. As a result, our choice of algorithm was DCRC for performing field trials.
The field trials were performed on a 200 $\textrm{km}^2$ region on Lake Murray, SC, USA. 
During the multi-robot coverage in a few instances, two vehicles came too close to each other. 

We are currently investigating an automated arbitration mechanism following the rules of the sea~\cite{Rules} to avoid collisions. Furthermore, a camera system is being developed to provide situational awareness of the  surroundings during operations. In general, the multi-robot coverage problem has several directions of interest, in particular taking into account the robustness of the proposed methods in real world.





\section*{ACKNOWLEDGMENT}
This work was made possible through the generous support of  National Science Foundation
grants (NSF 1513203, 1637876). The authors are grateful to Perouz Taslakian for invaluable advisement and Sharone Bukhsbaum, Christopher McKinney, George Sophocleous, and Anthony Grueninger for their assistance in designing and building the ASVs. 

\bibliographystyle{IEEEtran}
\bibliography{./IEEEabrv,refs}


\end{document}